\pdfoutput=1

\documentclass[11pt]{article}

\usepackage[dvipsnames]{xcolor}
\usepackage{paralist}
\usepackage{hyperref}
\usepackage[preprint]{acl}

\usepackage{times}
\usepackage{latexsym}
\usepackage{url}

\usepackage{todonotes}
\usepackage{amsmath}
\usepackage{tablefootnote}
\usepackage{placeins}  

\usepackage[T1]{fontenc}

\usepackage[utf8]{inputenc}

\usepackage{microtype}

\usepackage{inconsolata}

\usepackage{soul}

\usepackage{multirow}
\usepackage{colortbl} 
\usepackage{array}
\usepackage{arydshln} 
\usepackage{makecell} 
\usepackage{pgfmath} 
\usepackage{xcolor}
\usepackage[dvipsnames]{xcolor}

\definecolor{amazon-color}{RGB}{102,194,165}
\definecolor{dadjokes-color}{RGB}{252,141,98}
\definecolor{headlines-color}{RGB}{141,160,203}
\definecolor{oneliners-color}{RGB}{231,138,195}

\newcommand{\colorcell}[1]{%
    \ifnum#1<50
        \cellcolor[HTML]{F8696B}#1
    \else\ifnum#1<53
        \cellcolor[HTML]{F8766D}#1
    \else\ifnum#1<55
        \cellcolor[HTML]{F98370}#1
    \else\ifnum#1<58
        \cellcolor[HTML]{FA9072}#1
    \else\ifnum#1<60
        \cellcolor[HTML]{FA9D75}#1
    \else\ifnum#1<63
        \cellcolor[HTML]{FBAA77}#1
    \else\ifnum#1<65
        \cellcolor[HTML]{FCB77A}#1
    \else\ifnum#1<68
        \cellcolor[HTML]{FCC47C}#1
    \else\ifnum#1<70
        \cellcolor[HTML]{FDD17F}#1
    \else\ifnum#1<73
        \cellcolor[HTML]{FEDE81}#1
    \else\ifnum#1<75
        \cellcolor[HTML]{FFEB84}#1
    \else\ifnum#1<78
        \cellcolor[HTML]{F0E784}#1
    \else\ifnum#1<80
        \cellcolor[HTML]{E0E383}#1
    \else\ifnum#1<83
        \cellcolor[HTML]{D1DE82}#1
    \else\ifnum#1<85
        \cellcolor[HTML]{C1DA81}#1
    \else\ifnum#1<88
        \cellcolor[HTML]{B1D580}#1
    \else\ifnum#1<90
        \cellcolor[HTML]{A2D07F}#1
    \else\ifnum#1<93
        \cellcolor[HTML]{92CC7E}#1
    \else\ifnum#1<95
        \cellcolor[HTML]{83C77D}#1
    \else\ifnum#1<98
        \cellcolor[HTML]{73C37C}#1
    \else
        \cellcolor[HTML]{63BE7B}#1
    \fi\fi\fi\fi\fi\fi\fi\fi\fi\fi\fi\fi\fi\fi\fi\fi\fi\fi\fi\fi
}

\usepackage{graphicx}
\usepackage{color}

\newcommand{\mnote}[1]{\textcolor{orange}{\{ #1 -- Mor\}}}

\newcommand{\selfnote}[1]{\textcolor{magenta}{\{ #1 \}}}
\newcommand{\remove}[1]{}
\newcommand{\xhdr}[1]{\noindent {\bf #1.} }

\title{One Joke to Rule them All? 
On the (Im)possibility of Generalizing Humor
}

\author{
  Mor Turgeman$^{1}$ \quad Chen Shani$^{2}$ \quad Dafna Shahaf$^{1}$ \\
  $^{1}$The Hebrew University of Jerusalem \quad $^{2}$Stanford University \\
  \texttt{mortur@cs.huji.ac.il, cshani@stanford.edu, dshahaf@cs.huji.ac.il}
}

\begin{document}
\maketitle
\begin{abstract}
Humor is a broad and complex form of communication that remains challenging for machines. Despite its broadness, most existing research on computational humor traditionally focused on modeling a specific type of humor. In this work, we wish to understand whether competence on one or more specific humor tasks confers any ability to transfer to novel, unseen types; in other words, is this fragmentation inevitable?
This question is especially timely as new humor types continuously emerge in online and social media contexts (e.g., memes, anti-humor, AI fails). If Large Language Models (LLMs) are to keep up with this evolving landscape, they must be able to generalize across humor types by capturing deeper, transferable mechanisms. 
To investigate this, we conduct a series of transfer learning experiments across four datasets, representing different humor tasks. We train LLMs under varied diversity settings (1-3 datasets in training, testing on a novel task). 
Experiments reveal that models are capable of some transfer, and can reach up to 75\% accuracy on unseen datasets;
training on diverse sources improves transferability (1.88-4.05\%) with minimal-to-no drop in in-domain performance. Further analysis suggests relations between humor types, with Dad Jokes surprisingly emerging as the best enabler of transfer (but is difficult to transfer to).
We release data and code\footnote{\url{https://github.com/morturr/HumorTransferLearning.git}}.
\end{abstract}

\section{Introduction}


Humor spans a wide range of styles and mechanisms, from puns and sarcasm to absurdity and satire \citep{attardo2024linguistic, raskin1979semantic}, many of which involve linguistic play, pragmatic inference, or violations of logical expectations \citep{suls1972two, attardo2000irony}. 
It is subjective and culturally dependent \citep{attardo2024linguistic, martin2018psychology}, making the detection, generation, and explanation of humor hard for humans and machines \citep{shafiei2025not, loakman2025comparingapplesorangesdataset, horvitz2024gettinghumorcraftinghumor}.
Despite the broad and diverse nature of humor, much of existing work on computational humor has focused on narrow, specialized tasks such as detecting humor in internet memes \citep{kumari2024let}, knock-knock jokes \citep{taylor2004computationally}, puns \citep{xu2024good, miller-etal-2017-semeval, cocchieri-etal-2025-call}, cartoons \citep{shahaf2015inside} or even ``That's what she said'' jokes \citep{kiddon2011s},
but relatively little attention has been paid to learning the general phenomenon of humor.

In this work, we are interested in {\bf whether competence on one or more specific humor tasks confers any ability to transfer to novel, unseen types}. That is, we wish to understand whether splitting humor into subproblems is a necessary design choice or a historical artifact. This is especially important as new humor variants emerge over time; should we expect LLMs to understand them without further training, as humans often do?

Researchers from neuroscience and psychology have studied whether skill in one type of humor category aids another \emph{in humans}, and the evidence is mixed:
Findings suggest shared mechanisms (e.g., incongruity resolution) that 
provide some common ground across joke types and may enable partial transfer, but also specialized skills (language ambiguity, theory-of-mind, cultural knowledge) that are type-specific and create ``transfer costs'' \cite{dai2017resolve,farkas2021humor}  (see Section~\ref{sec:related_work}).

In NLP, several studies have explored transfer between different types of humor or languages \citep{arora2022transfer, baranov-etal-2023-told, wang2020unified}. There is some evidence that multi-category training helps humor detection, but these works did not test on humor types that were not a part of training, making it difficult to assess true generalization to novel types of humor. Moreover, they rarely
discuss the relations between humor types.



In this work we experiment with four humor-related datasets, representing different types of humor. We selected two advanced models, specifically chosen because they demonstrated poor performance on these datasets in a zero-shot setting; this allowed us to evaluate their potential for improvement through transfer. Our contributions are:

\begin{compactitem}
    \item We present the first systematic evaluation of humor transfer learning across multiple humor types using LLMs.
    \item We analyzed the models' performance in single and multi-task {humor binary classification} settings. We find that  models are capable of
    some transfer, with one model achieving 75\% accuracy on an unseen dataset.  \textbf{Training on diverse data encompassing multiple humor types enhances transfer to unseen types}.  
    In-domain performance remains relatively stable as the training set diversity increases, even as the number of training examples from the domain decreases significantly.
    \item We discover that certain humor types (e.g., Dad Jokes) more effectively enable transfer to others, suggesting latent structural relations.
    \item We propose a framework for studying humor transfer, including developing a method to generate negative (non-funny) examples for a dataset that includes only positive (funny) ones. We make our code and data public\footnotemark[1].


\end{compactitem}

\section{Research Questions}
\label{sec:rqs}
We investigate the capacity of LLMs to perform transfer learning across different types of humor. 
We address the following research questions (RQs): 

\begin{compactitem}
  \item {\bf RQ1}: Do LLMs have the capability for humor transfer learning? Can they learn some type(s) of humor and generalize to new humor types?

  \item {\bf RQ2}: Between which types of humor is transfer most effective? 

  \item {\bf RQ3}: How does data diversity influence humor transfer learning? Does training on more diverse datasets (e.g., containing multiple humor types) enhance generalization? 

\end{compactitem}




\begin{table*}[t]
  \centering
  \begin{tabular}{p{2cm} p{2.5cm} p{5cm}p{5cm}}
    \hline
    \textbf{Name} & \textbf{Text Length Mean \& Std} & \textbf{Positive Example}
    & \textbf{Negative Example}\\
    \hline\hline
    Amazon Questions & $143.80 \pm 58.64$ $\text{*} 60.41 \pm 37.21$
    & PEREGRINE Banana Saver Yellow | I have a problem with wolves where I live. Will this carrier protect bananas from wolves?  
    &  Two-Wheel Smart Scooter Self Balancing Unicycle Electric Scooter Electric Unicycle Smart Wheel | Do you ship by ups or fedex? \\\hline
    
    Reddit Dad Jokes & $ 86.10 \pm 26.25 $
    & I went to a bookstore and asked where the self-help section was The clerk said that if she told me, it would defeat the purpose. 
    & Did you hear of the Librarian who became unwell while reading a book? She had to take a sick leave. \\\hline
    Sarcasm Headlines & $ 62.45 \pm 21.10 $
    & new york introduces shoe-sharing program for city's pedestrians
    & stars with gray hair prove getting older isn't all that bad \\\hline
    One Liners & $ 60.66 \pm 18.44 $
    & Couldn’t afford to fix my brakes, so I made my horn
louder.
    & Fear a silent man. He has lips like a drum. \\\hline\hline
  \end{tabular}
  \caption{Properties of the datasets used in our experiments. For each dataset, we report the mean and standard deviation of text length, along with example sentences from the positive (humorous) and negative (non-humorous) classes. \textit{For the Amazon Questions dataset, we report both the length of the full input (product name + question) and—marked with an asterisk—the length of the user question alone (i.e., the humorous content): $60.41 \pm 37.21$}.
}
  \label{tab:dataset examples}
\end{table*}

\section{Data}

\label{sec:data}
\remove{
\begin{table*}[t]
  \centering
  \begin{tabular}{p{3cm} p{3cm} p{2cm} p{2cm} p{1.5cm} p{1.5cm}}
    \hline\hline
    \textbf{Name} & \textbf{Source} & \textbf{Original Size}
    & \textbf{Sampled (\%)} & \textbf{Positive Selection} & \textbf{Negative Selection}\\\hline
    \hline
    Amazon Questions & \citet{Ziser2020HumorDI} & 19K
    & 33\% & - & - \\\hline
    Reddit Dad Jokes & Kaggle \citep{reddit_dadjokes_dataset} & 216K 
    & 3\% & score >= 20 & generated by GPT-4 Turbo \\\hline
    Sarcasm Headlines & \citep{Misra2023SarcasmDU} & 28K
    & 22\% & - & - \\\hline
    One Liners & \citet{Mihalcea2005MakingCL} & 32K
    & 19\% & - & - \\
    \hline\hline
  \end{tabular}
  \caption{Overview of the datasets used in our experiments, including sources, original sizes, sampling rates, and criteria for selecting positive and negative examples. \mnote{remove here the columns of pos\&neg selection? because we removed yelp, it is relevant only to dad jokes}}
  \label{tab:dataset info}
\end{table*}
}

To answer these RQs, we experiment with 4 humor datasets, each focusing on a distinct humor type. These datasets vary in style, domain, and structure, offering a diverse testbed for generalization.\footnote{While these datasets may contain examples spanning multiple humor genres, they provide a foundation for analyzing relationships and transferability across humor types.}
%
%

\subsection{Dataset Descriptions}

Table~\ref{tab:dataset examples} provides representative examples from both the humorous and non-humorous classes of all the datasets, as well as mean text length.

\noindent {\textbf{Amazon:}} 19K records, each containing a product name paired with a user-submitted question, annotated for humor by humans \citep{Ziser2020HumorDI}.
  
\noindent {\textbf{One Liners:}} 32K one-liner sentences \citep{Mihalcea2005MakingCL}. 
Humorous examples were collected using an algorithm designed to harvest funny one-liners. 
Non-humorous examples were sourced from news headlines, proverbs, and sentences from the British National Corpus.
 
\noindent{\textbf{Sarcasm Headlines:}} 28K headlines consisting of both real news headlines and sarcastic ones from \textit{The Onion}, a satirical news outlet \citep{Misra2023SarcasmDU}. 
  
\noindent{\textbf{Reddit Dad Jokes:}} Reddit posts collected from the \texttt{r/dadjokes} subreddit \citep{reddit_dadjokes_dataset}. As the original dataset only includes positive (i.e., humorous) examples, we generated negative (non-humorous) samples to enable binary classification. 

{For the positive class, we selected high-confidence positive samples (Reddit score $\geq 20$).} To create negative examples that closely match the positive class in content, writing style, and semantics, we used GPT-4 Turbo \citep{openai2023devday} in a few-shot setup. For the other samples (lower Reddit score), we asked GPT to minimally modify each joke, preserving style and content but removing the humorous element. E.g., given ``Why can't milk cartons walk? Because they lactose,'' the generated negative was ``Why can't milk containers move? They lack appendages.'' See prompt in Appendix~\ref{app:gpt-prompt}. 


\remove{\begin{figure}
  \includegraphics[width=\columnwidth]{llm_prompt_big (1).pdf}
  \caption[width=0.7\textwidth]{Generating the negative class of Reddit Dad Jokes using GPT-4-Turbo. A joke from the positive class is provided as input, and the model outputs a non-humorous sentence that retains the original joke’s content and writing style.}
  \label{fig:gpt4-output}
\end{figure}}

To assess generation quality, we manually reviewed 3,000 outputs. Only 2.63\% did not maintain the style or content (typically due to the LLM summarizing the unfunny joke). Importantly, we found no examples where the punchline was retained.


To ensure balanced text length distributions between both classes {of Dad Jokes}, we paired each negative example with the closest-length positive example, ensuring no duplicates. 

\subsection{Dataset Properties and Preprocessing}
\label{sec:dataset-properties}
\begin{figure*}[t!]
\begin{center}
  \includegraphics[width=1\textwidth]{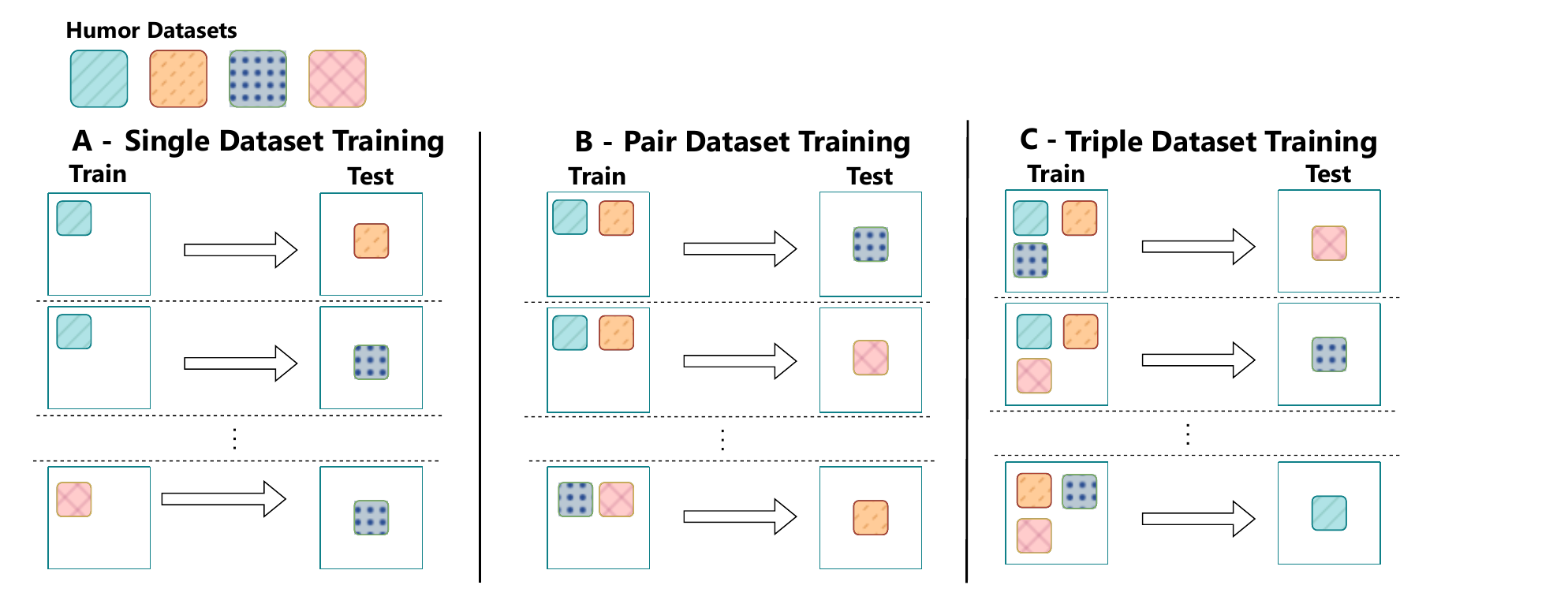}
  \caption[width=\textwidth]{Overview of the experimental setups.
\textbf{(A) Single Dataset Training:} Train on one dataset and test on each of the other datasets.
\textbf{(B) Double Dataset Training:} Train on two of datasets and test on each of the remaining three.
\textbf{(C) Triple Dataset Training:} Train on three datasets and test on the held-out forth dataset.}
  \label{fig:experiments}
\end{center}
\end{figure*}

\xhdr{Humor Styles}  
The datasets span a variety of humor styles: Questions in the Amazon dataset are often sarcastic or ironic, Reddit Dad Jokes typically follow a short-story format, Sarcastic News Headlines feature more sophisticated or absurd humor that is often borders on non-sequitur, and One-Liners and Dad Jokes both include brief, standalone jokes, frequently employing puns or wordplay.

\xhdr{Data Partitioning}  
For fair cross-dataset comparisons, we randomly downsampled all datasets to 6,250 examples. Each dataset was balanced between positive and negative classes. We used an 80\%/2\%/18\% training/validation/test split (to balance evaluation reliability with efficiency, given the large number of trained models and evaluations).


In transfer experiments from multiple datasets, we constructed the training set by sampling equally from each dataset, ensuring class balance and maintaining a consistent training size of 5,000 (2,500 positive and 2,500 negative datapoints).

\remove{
\xhdr{Expectations before the Experiments}
We expected 
Dad Jokes and One-Liners to exhibit transferability due to their shared reliance on puns and short-form humor. Headlines also structurally resemble One-Liners, so we expected partial transfer.
}


\section{Experiments}

We design a series of experiments to answer our RQs (do LLMs transfer humor learning, the effect of data diversity, and humor type; see Section \ref{sec:rqs}). Figure~\ref{fig:experiments} depicts our three experimental setups.

\begin{compactitem}
    \item \textbf{Single Dataset Training:}  
    Examines whether basic transfer occurs between datasets (RQ1) and whether certain datasets are more similar to others (RQ2). Models were fine-tuned on each dataset and evaluated on all datasets. 

    \item \textbf{Double Dataset Training:}  
    Explores multi-task learning by examining how training on two humor styles affects both in-domain performance and generalization to other styles (RQ3), and further investigates the relationships between different humor types (RQ2). Models were fine-tuned on each pair of datasets and evaluated on all datasets.  

    \item \textbf{Triple Dataset Training:}  
     Examines the effect of training in the most diverse data setting to evaluate how data diversity impacts transferability (RQ3), and to identify which humor types are most generalizable from others (RQ2). Models were trained on three datasets and evaluated on all datasets.
\end{compactitem}
    
\xhdr{Models} For all experiments, we tested both LLaMA-2-7B \citep{touvron2023llama} and Mistral-7B \citep{jiang2023mistral7b}. We selected these models for three main reasons: (1) they are comparable in size, ensuring a fair comparison in terms of capacity; (2) they belong to different model families, allowing us to assess whether transferability trends are consistent across architectures; (3) both models are widely used and have demonstrated strong performance on various tasks, making them suitable candidates for humor-related evaluation, and (4) both models performed near guess-level (40\%-56\%) in a zero-shot manner when tested on our validation datasets, leaving room for improvement and transfer (see details in Appendix~\ref{app:zero-shot}). 
    
We applied instruction fine-tuning to the models and used a prompt describing the task for each training sample (see Appendix \ref{app:instruction-finetune-prompt}).
Full training and evaluation details are provided in Appendix~\ref{app:training-details}.

\begin{table*}
    \centering
    \begin{tabular}{cccccc}
        \hline
        \multirow{2}{*}{\textbf{Train Dataset}} &
        \multirow{2}{*}{\textbf{Model}} &
        \multicolumn{4}{c}{\textbf{Test Dataset}} \\ \cline{3-6} &
                                            & \textbf{\textcolor{amazon-color}{Amazon}} & \textbf{\textcolor{dadjokes-color}{Dad Jokes}} & \textbf{\textcolor{headlines-color}{Headlines}} & \textbf{\textcolor{oneliners-color}{One Liners}} \\ \hline
        \multirow{2}{*}{\textbf{\textcolor{amazon-color}{Amazon}}} & LLaMA-2 &
        \colorcell{88} & \colorcell{65} & \colorcell{65} & \colorcell{61} 
        
        \\\cline{2-6} & Mistral &
        \colorcell{91} & \colorcell{62} & \colorcell{75} & \colorcell{72}
        \\\Xhline{1pt}

        \multirow{2}{*}{\textbf{\textcolor{dadjokes-color}{Dad Jokes}}} & LLaMA-2 &
        \colorcell{63} & \colorcell{93} & \colorcell{59} & \colorcell{70} 

        \\\cline{2-6} & Mistral &
        \colorcell{69} & \colorcell{94} & \colorcell{68} & \colorcell{71} 
        \\\Xhline{1pt}

        \multirow{2}{*}{\textbf{\textcolor{headlines-color}{Headlines}}} & LLaMA-2 &
        \colorcell{58} & \colorcell{57} & \colorcell{90} & \colorcell{65}

        \\\cline{2-6} & Mistral &
        \colorcell{65} & \colorcell{56} & \colorcell{97} & \colorcell{62}
        \\\Xhline{1pt}

        \multirow{2}{*}{\textbf{\textcolor{oneliners-color}{One Liners}}} & LLaMA-2 &
        \colorcell{62} & \colorcell{62} & \colorcell{54} & \colorcell{87} 

        \\\cline{2-6} & Mistral & 
        \colorcell{64} & \colorcell{51} & \colorcell{67} & \colorcell{95} 

        \\\hline
        
    \end{tabular}
    \caption{\textbf{[Partial transfer between humor datasets.]} \textbf{Single Dataset Training:} Accuracy scores (0–100), averaged over four training seeds. Models trained on a single dataset and evaluated on all. Diagonal values reflect in-domain performance, showing that both models effectively learn their respective datasets. Off-diagonal values capture transfer performance, revealing asymmetric transfer patterns. For example, Dad Jokes transfers well to One Liners (70-71\%), but not vice versa (51–62\%). Mistral consistently shows stronger transfer than LLaMA-2, especially when trained on Amazon and Dad Jokes. See Appendix Table~\ref{tab:std-table} for standard deviations.}

    \label{tab:single dataset results}
\end{table*}

\begin{table*}

    \centering
    \begin{tabular}{cccccc}
        \hline
        \multirow{2}{*}{\textbf{Two Datasets}} &
        \multirow{2}{*}{\textbf{Model}} &
        \multicolumn{4}{c}{\textbf{Test Dataset}} \\ \cline{3-6}
                                            & & \textbf{\textcolor{amazon-color}{Amazon}} & \textbf{\textcolor{dadjokes-color}{Dad Jokes}} & \textbf{\textcolor{headlines-color}{Headlines}} & \textbf{\textcolor{oneliners-color}{One Liners}} \\ \Xhline{1pt}
                                            
        \multirow{2}{*}{\textbf{\textcolor{amazon-color}{Amazon} + \textcolor{dadjokes-color}{Dad Jokes}}} & LLaMA-2 &
        \colorcell{82} & \colorcell{90} & \colorcell{67} & \colorcell{69}
        
        \\\cline{2-6} & Mistral 
        & \colorcell{89} & \colorcell{95} & \colorcell{74} & \colorcell{74} 
        \\\Xhline{1pt}

        \multirow{2}{*}{\textbf{\textcolor{amazon-color}{Amazon} + \textcolor{headlines-color}{Headlines}}} & LLaMA-2 &
        \colorcell{82} & \colorcell{62} & \colorcell{90} & \colorcell{69}
        
        \\\cline{2-6} & Mistral &
        \colorcell{89} & \colorcell{67} & \colorcell{96} & \colorcell{67}
        \\ \Xhline{1pt}

        \multirow{2}{*}{\textbf{\textcolor{dadjokes-color}{Dad Jokes} + \textcolor{headlines-color}{Headlines}}} & LLaMA-2 &
        \colorcell{63} & \colorcell{89} & \colorcell{92} & \colorcell{71}
        
        \\\cline{2-6} & Mistral &
        \colorcell{71} & \colorcell{91} & \colorcell{95} & \colorcell{70} 
        \\ \Xhline{1pt}

        \multirow{2}{*}{\textbf{\textcolor{dadjokes-color}{Dad Jokes} + \textcolor{oneliners-color}{One Liners}}} & LLaMA-2 &
        \colorcell{74} & \colorcell{90} & \colorcell{65} & \colorcell{91}

        \\\cline{2-6} & Mistral &
        \colorcell{66} & \colorcell{95} & \colorcell{70} & \colorcell{94}
        \\ \Xhline{1pt}

        \multirow{2}{*}{\textbf{\textcolor{headlines-color}{Headlines} + \textcolor{oneliners-color}{One Liners}}} & LLaMA-2 &
        \colorcell{63} & \colorcell{55} & \colorcell{93} & \colorcell{92}
        
        \\\cline{2-6} & Mistral
        & \colorcell{68} & \colorcell{52} & \colorcell{97} & \colorcell{94}  
        \\\Xhline{1pt}

        \multirow{2}{*}{\textbf{\textcolor{oneliners-color}{One Liners} + \textcolor{amazon-color}{Amazon}}} & LLaMA-2 &
        \colorcell{86} & \colorcell{56} & \colorcell{65} & \colorcell{91}
        
        \\\cline{2-6} & Mistral &
        \colorcell{90} & \colorcell{53} & \colorcell{74} & \colorcell{94} 
        \\ \Xhline{1pt}

    \end{tabular}
    \caption{\textbf{[Amazon + Dad Jokes generalizes best.]} \textbf{Double Dataset Training:} Accuracy scores (0–100), averaged over four seeds. Models were trained on two datasets. Amazon + Dad Jokes yields the strongest transfer (74\% Mistral; 67–69\% LLaMA-2), while Headlines + One Liners yields the weakest transfer (52–55\% on Dad Jokes; 63–68\% on Amazon). Mistral outperforms LLaMA-2 in most cases. See Appendix Table~\ref{tab:std-table} for standard deviations.}

    \label{tab:full pair results}
\end{table*}

\begin{table*}
\centering
\begin{tabular}{cccccc}
    \hline
    \multirow{2}{*}{\textbf{Left Out Dataset}} &
    \multirow{2}{*}{\textbf{Model}} &
    \multicolumn{4}{c}{\textbf{Test Dataset}} \\ \cline{3-6} &
                                        & \textbf{\textcolor{amazon-color}{Amazon}} & \textbf{\textcolor{dadjokes-color}{Dad Jokes}} & \textbf{\textcolor{headlines-color}{Headlines}} & \textbf{\textcolor{oneliners-color}{One Liners}} \\ \Xhline{1pt}
                                        
    \multirow{2}{*}{\textbf{\textcolor{amazon-color}{Amazon}}} & LLaMA-2 &
    \colorcell{69} & \colorcell{88} & \colorcell{89} & \colorcell{88} 
    
    \\\cline{2-6} & Mistral &
    \colorcell{66} & \colorcell{94} & \colorcell{96} & \colorcell{94}
    \\\Xhline{1pt}

    \multirow{2}{*}{\textbf{\textcolor{dadjokes-color}{Dad Jokes}}} & LLaMA-2 &
    \colorcell{84} & \colorcell{57} & \colorcell{90} & \colorcell{87}
    
    \\\cline{2-6} & Mistral &
    \colorcell{90} & \colorcell{55} & \colorcell{96} & \colorcell{94}
    \\ \Xhline{1pt}

    \multirow{2}{*}{\textbf{\textcolor{headlines-color}{Headlines}}} & LLaMA-2 &
    \colorcell{86} & \colorcell{89} & \colorcell{68} & \colorcell{88} 
    
    \\\cline{2-6} & Mistral &
    \colorcell{90} & \colorcell{95} & \colorcell{73} & \colorcell{93} 
    \\ \Xhline{1pt}

    \multirow{2}{*}{\textbf{\textcolor{oneliners-color}{One Liners}}} & LLaMA-2 &
    \colorcell{84} & \colorcell{89} & \colorcell{91} & \colorcell{69}

    \\\cline{2-6} & Mistral &
    \colorcell{90} & \colorcell{96} & \colorcell{96} & \colorcell{74}

    \\ \hline
\end{tabular}
\caption{\textbf{[Training on limited data preserves self accuracy.]} \textbf{Triple Dataset Training:} Accuracy scores (0–100), averaged over four seeds. Models were trained on three datasets (excluding the one listed in the “Left Out Dataset” column), using 33\% of each. Strong seen-dataset accuracy shows robust learning; partial transfer is observed on the left-out dataset (e.g., Mistral reaches 73\% on Headlines, 74\% on One Liners). See Appendix Table~\ref{tab:std-table} for STDs.}

\label{tab:all loo results}
\end{table*}

\section{Results and Analysis}
We now present and analyze our experimental results, addressing each RQ in turn. Accuracy scores are summarized in Tables~\ref{tab:single dataset results}, \ref{tab:full pair results}, and \ref{tab:all loo results} (deviations across training seeds are in Appendix Table~\ref{tab:std-table}).
    \subsection{RQ1: Transfer Humor Capability}
    \xhdr{LLMs can transfer humor knowledge across datasets, but success varies by model and humor type}
    To assess whether LLMs can perform humor transfer learning, we examine the results from the Single Dataset Training (Table~\ref{tab:single dataset results}). Both LLMs are able to learn the humor style they were trained on (in-domain), but they differ in their ability to generalize to unfamiliar humor types (transfer).

    LLaMA-2 shows weaker performance overall, with in-domain accuracy averaging 4.75\% lower than Mistral's. Its cross-dataset performance is $\sim$60\% in most cases, exhibiting some transfer.
    Mistral demonstrates better transfer, reaching 67–75\% accuracy on several target datasets, particularly when trained on Amazon or Dad Jokes.

    \subsection{RQ2: Linking Humor Types}
    \label{sec:dataset-relationship}
    \xhdr{Humor types differ in transferability, forming a hierarchy from highly generalizable styles to more complex ones that require focused learning}
    We investigate which datasets enable the most effective transfer across three experimental setups. 
    
    In the single-dataset-training experiment (Table \ref{tab:single dataset results}) we focus on Mistral, given its superior performance. Training on Amazon leads to relatively high transfer accuracy on Headlines (75\%) and One Liners (72\%), whereas the reverse direction yields lower performance (64–65\%). This asymmetry suggests that Amazon may support broader generalization, potentially due to its diverse, user-generated content covering a wide range of humor styles and topics. In contrast, Headlines and One Liners are more structurally constrained and stylistically homogeneous, limiting their transferability.
    
    Dad Jokes shows the most asymmetric pattern: training on it yields strong transfer accuracy (68–71\%), but models trained on other datasets perform poorly when evaluated on Dad Jokes (51–62\%). 
    We note that its structure often involves multi-sentence narratives, puns, irony, and cultural references, which are not easily captured by shorter or more templatic humor styles.
    
    Finally, One Liners and Headlines show relatively strong bidirectional transfer, likely due to their shared brevity and stylistic similarity. Both rely on compact, punchline-driven formats and often draw from news-style or everyday language, which may facilitate mutual generalization.

    These findings suggest that \textbf{more general, diverse types of humor support broader transfer}, while narrowly framed or format-constrained types exhibit limited generalization.
    
    We analyze the pairwise training results in Table~\ref{tab:full pair results}, computing the average transfer accuracy for each target dataset by averaging model performance across all training pairs that exclude it.
    Both LLMs exhibit similar trends across most datasets, mirroring the earlier conclusion that \textbf{Dad Jokes is the most complex humor style, whereas One Liners and Headlines are more learnable from other humor types} (Table~\ref{tab:full pair results}). 
    For every target dataset, the strongest transfer is obtained when training pairs include Dad Jokes. 
    This strengthens the conclusion from single-dataset-training that Dad Jokes captures the broadest range of humor, followed closely by Amazon, thereby enabling wider generalization to unseen styles.

    Conversely, when Dad Jokes is the target, both LLMs perform best when trained on Amazon + Headlines (62–67\%), outperforming combinations containing One Liners (52–56\%). This mirrors the weak One Liners → Dad Jokes transfer observed in the single-source setting and further illustrates the strong asymmetry between these humor types.
    
    We now examine transfer performance in the triple-dataset experiment, where each dataset is held out once for evaluation (Table~\ref{tab:all loo results}). As in the previous experiments, both models struggle to transfer to Dad Jokes, with an average accuracy of 56\%. In contrast, Headlines and One Liners show the strongest transfer results, averaging 70.5\% and 71.5\% respectively, followed by Amazon, which demonstrates slightly weaker transfer with an average of 67.5\%. These findings are consistent with the trends observed in the single- and double experiments, where Dad Jokes emerged as the most difficult humor type for transfer learning, and One Liners and Headlines were the most receptive to transfer from different humor styles.
    
    Taken together, the experiments \textbf{reveal a hierarchy of humor complexity and transferability: Dad Jokes enables strong transfer to all datasets but remains difficult to generalize to}. Amazon occupies a middle ground, benefiting from Dad Jokes while transferring reasonably well to simpler styles. \textbf{Headlines and One Liners are the most generalizable targets}, but offer comparatively less utility when used as sources for transfer.

        \begin{figure}
      \includegraphics[width=\columnwidth]{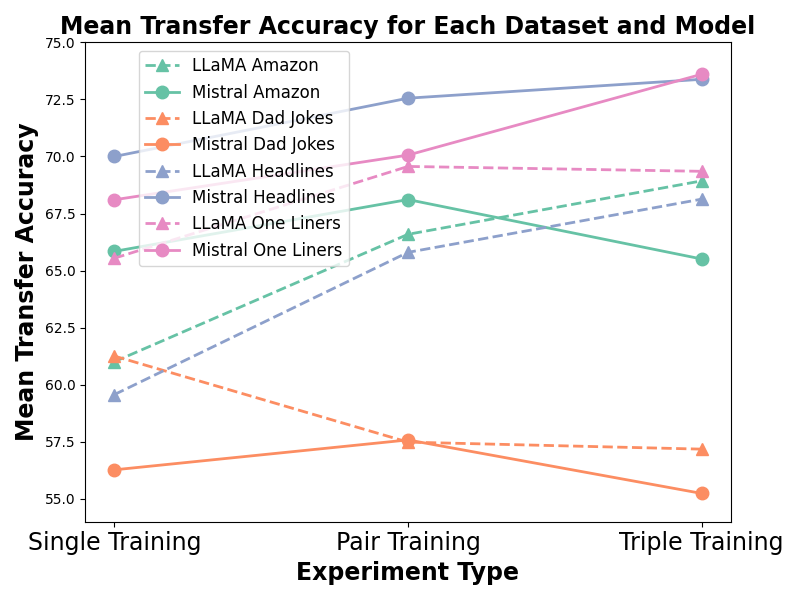}
      \caption{\textbf{[Increasing the training data diversity improves transfer.] Comparing transfer across the experiments.}
    Mistral results are shown with solid lines, LLaMA-2 with dotted lines. Colors represent different test datasets. The x-axis indicates the experiment type, and the y-axis shows the mean transfer accuracy. In general, more diverse training data leads to better transfer than single-dataset training. LLaMA-2 shows consistent improvement across experiments, except on Dad Jokes. Notably, Dad Jokes is the only dataset that performs worse with increased data diversity.}

      \label{fig:transfer_comparison}
    \end{figure}
    
    \subsection{RQ3: Impact of Data Diversity}
    We now explore how the diversity of humor \textit{training} data affects model performance. Specifically, whether exposure to multiple humor styles improves generalization across domains, and whether it depends on humor type or evaluation setting. We note that we discuss the average case in our experiments; of course, adding a dataset  that is similar to the target  could affect transfer dramatically. 

    We first assess the overall impact of data diversity on humor transfer learning. Next, we investigate how different humor types respond to diversity during training, identifying styles that benefit more or less from multi-source input. We then evaluate how data diversity influences in-domain accuracy (that is, performance on humor styles included in training). Finally, we compare the effects of diversity between Mistral and LLaMA-2 to understand whether model architecture or pretraining background modulates these trends.

    \subsubsection{Impact of Data Diversity on Transfer}
    \xhdr{Greater training diversity improves humor transfer, but with diminishing returns}
    To investigate how training data diversity affects humor transferability, we compare our three experiments: (A) single-dataset training (no diversity), (B) double-training (moderate diversity), and (C) triple-training (high diversity). For each setup, we evaluate on a held-out dataset and average performance across runs that exclude the evaluated dataset from training. Figure~\ref{fig:transfer_comparison} depicts transfer performance across the three experiments (non-aggregated results are in Table~\ref{tab:transfer diffs} in Appendix \ref{app:transfer-diffs}).

    
    For LLaMA-2, we observe a consistent improvement in transfer accuracy with increasing diversity. Moving from single- to double-dataset training yields an average gain of 3.02 percentage points across target datasets, with further gains of 1.04 points from double to triple. The only exception is Dad Jokes, which shows the opposite trend, arguably due to its relative high diversity.
    
    Mistral follows a similar trend from single-to-double-dataset training, improving by 2.02 points on average. However, it plateaus in triple-training, with a slight drop of 0.14 points compared to double, though still 1.88 higher than single-training.
    

    Training on {diverse humor sources consistently improves models' generalization to unseen humor types}. The largest gains come from moving beyond single-dataset training; returns begin to diminish as more datasets are added, suggesting that \textbf{moderate diversity may be nearly as effective as maximal diversity for cross-domain humor transfer}.

    \subsubsection{Transfer Between Distinct Humor Types}
    \xhdr{Data diversity yields larger gains for structurally simpler humor types}
    We now focus on humor transfer across \textit{humor types}. Headlines and One Liners consistently benefit from increased diversity, with both LLMs showing improved accuracy from single- to double- to triple-dataset training. A minor exception occurs for LLaMA on One Liners, where double- slightly outperforms triple-training, though both exceed single baseline. 
    
    For Amazon, LLaMA's accuracy improves substantially when increasing diversity. In contrast, for Mistral, double-dataset-training yields the highest accuracy, while the triple performing slightly worse than single-dataset baseline. 
    Dad Jokes displays a decline in performance as data diversity increases. 
    
    These findings follow the datasets complexities we observed in Section \ref{sec:dataset-relationship} suggest that \textbf{increasing training data diversity is more beneficial for generalizing to simpler humor types}. However, for complex humor types, increased diversity may not be as helpful, as the models may need to internalize the unique structure and nuances of the genre.


    \subsubsection{Data Diversity \& In-Domain Accuracy}
    \xhdr{Less in-domain training data leads to only a small drop in in-domain performance}
    We evaluate the impact of reduced dataset specific training data on in-domain performance. For each dataset, we compare its single-dataset training accuracy to the average accuracy of the three models trained on it within a triple-dataset setup (Tables~\ref{tab:single dataset results} \& \ref{tab:all loo results}).

    In most cases, in-domain accuracy decreases under the triple-training setup due to the reduced amount of in-domain data (just 33\% compared to single-dataset training). However, the average accuracy of Mistral drops by only 0.49 percentage points, while LLaMA-2 sees a slightly larger decrease of 1.76 points. These results suggest that both models are relatively \textbf{robust to reductions in in-domain data, with only minimal performance loss} even when training data is substantially diluted. In rare cases, the increased training diversity in the triple setup led to small performance gains.



    \subsection{Dataset Embedding Similarity}
    To better understand dataset relations, we computed pairwise cosine similarity between sentence embeddings from the training split, both within and across datasets (see Appendix~\ref{app:cosine_heatmaps}). We used Mistral, which produced more robust results. Surprisingly, the most complex datasets, Dad Jokes and Amazon, were also the most self-similar, while One Liners was the least. This contradicts our hypothesis that broader datasets would exhibit greater internal diversity. Notably, \textbf{higher cross-dataset similarity often corresponds to stronger observed transfer}.
    


\subsection{Prioritizing Diversity in Humor Learning} 
As our results show, increasing data diversity generally enhances humor transfer learning, echoing trends observed in other NLP tasks \citep{yu-etal-2022-data, wang2022generalizingunseendomainssurvey, rozen-etal-2019-diversify}. 
Reducing the amount of in-domain data to 33\% 
led to only a slight decrease in accuracy.
This could mean 
our datasets were large enough to retain performance despite downsampling, or that future humor applications should prioritize data diversity over size 
(similar to recent work highlighting the importance of diversity in synthetic datasets \citep{zhu2025mattersllmgenerateddatadiversity, long-etal-2024-llms, chen2024diversitysyntheticdataimpact}).

    \subsection{Comparing Mistral and LLaMA-2} 
    While both capture humor and exhibit transfer, they show distinct strengths. Notably, Mistral consistently outperforms LLaMA-2 in transfer settings, suggesting a superior ability to learn shared stylistic or structural features across humor types. 
    
    Despite differences in performance, both models exhibit consistent patterns: (1) Dad Jokes reliably support transfer to other datasets but remain hard to generalize to; (2) Headlines and One Liners are easy to generalize to but offer limited transfer benefit; (3) Amazon occupies a middle ground, both benefiting from and contributing to transfer; and (4) data diversity benefits simpler humor types, while complex styles require more focused exposure.

    The alignment across models supports the conclusion that humor transfer is governed by structured, humor-type-specific patterns. Still, performance gaps such as Mistral achieving nearly 70\% accuracy in settings where LLaMA-2 falls short, raise questions about whether the extent of transferability is intrinsic to humor or dependent on model architecture. These discrepancies highlight important directions for future research.

\section{Related Work}
\label{sec:related_work}

\paragraph{Humor Taxonomy.}
Linguistic and psychological theories provide rich taxonomies of humor. \citet{tsakona2017genres} distinguish humor types by contextual expectations, while \citet{beyond_a_joke} categorize forms like irony, puns, and allusions. The Humor Styles Questionnaire \citep{MARTIN200348} defines four psychological humor styles with distinct social functions. While these frameworks describe humor in depth, prior work has not systematically examined relationships between humor types or their potential for structural clustering or transfer.

\paragraph{Humor Transfer in Humans.}
Neuroimaging studies confirm that different humor types can recruit distinct brain systems, which implies limited transfer.
For instance, researchers found that complex semantic jokes activated temporal-lobe language areas, whereas sound-based puns engaged left frontal speech-processing regions \cite{martin2018psychology}.  \citet{dai2017resolve} showed that typical ``incongruity-resolution'' (resolvable) humor and absurd humor
(unresolvable) involve different neural paths in all
stages. 
One fMRI meta-analysis found that
humor comprehension engages broad language and reward circuits regardless of stimulus type,
but ToM-based humor specifically activates mentalizing areas \cite{farkas2021humor}. 
%
Developmental studies corroborate these results: young children grasp simple jokes earlier than ironic sarcasm \cite{angeleri2014development}, and people with strong verbal skills do better on semantic
jokes \cite{yankovitz2023relationship}. Together these findings suggest shared mechanisms that 
provide some common ground,
but also specialized skills
that are type-specific.

\paragraph{Computational Humor.}
Humor recognition is subjective, context- and culture-sensitive. \citet{kalloniatis2024computational} reviewed the complexity of humor datasets and models. 
Multimodal, cross-lingual, and application-oriented studies \citep{xie2023multimodal, shani2022alexa, shapira2023evaluating} further highlight the field's breadth.

\paragraph{Transfer Learning.}
From a machine learning perspective, our work builds on the foundation of transfer learning and multi-task learning (MTL). See \citet{zhuangtransferlearningsurvey,zhang2021surveymultitasklearning} for a broad survey.

\paragraph{Humor Transfer in LLMs.}
Prior MTL work on humor focused on joint training rather than transfer. \citet{arora2022transfer} used a shared-private model to capture general and type-specific humor features but did not test generalization to unseen types. \citet{baranov-etal-2023-told} explored transfer by training on multiple humor datasets and evaluating on overlapping subsets, finding that diversity aids generalization. In contrast, we are interested in transfer to entirely unseen datasets.
%
\citet{wang2020unified} tackled multilingual tasks but did not explore transfer across languages. 
{\citet{loakman2025comparingapplesorangesdataset} investigated the ability of LLMs to explain jokes across different humor types. They found that none of the tested LLMs are consistently capable of generating adequate explanations for all joke types. 
}

\section{Conclusions}
Humor is a complex and varied domain, yet not opaque to transfer-based learning. In this work, we asked whether competence on specific humor types enables generalization to novel styles. We found that transfer is possible but asymmetric: types like Dad Jokes support transfer but are hard to generalize to, while Headlines and One Liners are easier to predict but contribute little to transfer. These patterns reflect structural differences among humor types and align with cognitive theories distinguishing surface cues from deeper mechanisms.

Exposure to diverse humor types generally improves performance, particularly for simpler styles, though nuanced forms often require focused, style-specific training. While both LLaMA-2 and Mistral capture broad transfer patterns, Mistral consistently generalizes better. Interestingly, models retain strong in-domain performance even when trained on only 33\% of target data.

Future work should expand to more humor types and modalities (e.g., cartoons or internet memes) and explore different axes of transfer, such as multilingual or cross-culture settings, as well as seek
alignment between transfer patterns and findings and theories from cognition and neuroscience.
We hope this work inspires follow-up work that could further illuminate what makes humor transferable in machines and in minds.

\newpage

\section{Limitations}

This study has several limitations. First, we focus exclusively on short-form, English-language textual humor, excluding multimodal formats (e.g., memes, videos) and interactive contexts such as dialogue or conversational humor. Second, while each dataset serves as a stand-in for a particular humor style, these assignments are approximate and do not capture the full nuance or variability of humor genres. Moreover, individual datasets may reflect specific demographics or cultural biases (e.g., Reddit Dad Jokes is representative of a particular community). Third, our analysis is based on a limited set of datasets (four) and models (Mistral-7B and LLaMA-2-7B), which may constrain the generalizability of our findings to other humor domains, linguistic settings, or model families. Future work could address these limitations by incorporating a broader range of humor types, languages, and architectures to better capture the richness and diversity of humorous expression.

\section{Ethical considerations}
Some of the datasets used in this study were collected from publicly available internet sources and may contain offensive content. Humor, by nature, often challenges social norms, and internet discourse can occasionally reflect inappropriate or sensitive material. However, a brief examination of the datasets revealed no indications of content that exceeds the bounds of good taste. We used the datasets as-is to preserve their original characteristics, which are essential for analyzing humor in natural contexts.

The datasets do not contain personally identifiable information, with the exception of the Reddit Dad Jokes dataset, which may include usernames. We note that this information is public (on Reddit), and we did not make any use of it in our work.

All datasets were used in accordance with their respective terms of use and solely for academic research purposes.

\appendix

\bibliography{custom}

\newpage
\section{Model Zero-Shot Performance}
\label{app:zero-shot}

We conducted a zero-shot evaluation using both LLMs across the validation splits of all datasets. The models received the instruction prompt (excluding the actual response; see Appendix~\ref{app:instruction-finetune-prompt}) and produced outputs of either ``Funny'' or ``Not funny,'' indicating that they understood the task format. Accuracy ranged from 40\% to 56\%, which is about guess level (see full results in Table \ref{tab:zero shot results}). Thus, the transfer patterns observed in our experimental setup did not occur randomly but resulted from learning humor-specific information during training.

\begin{table*}
    \centering
    \begin{tabular}{ccccc}
        \hline
        \multirow{2}{*}{\textbf{Model}} &
        \multicolumn{4}{c}{\textbf{Test Dataset}} \\ \cline{2-5} 
                                            & \textbf{\textcolor{amazon-color}{Amazon}} & \textbf{\textcolor{dadjokes-color}{Dad Jokes}} & \textbf{\textcolor{headlines-color}{Headlines}} & \textbf{\textcolor{oneliners-color}{One Liners}} \\ \hline
         \textbf{LLaMA-2} &
        \colorcell{55} & \colorcell{56} & \colorcell{56} & \colorcell{49} 
        \\\Xhline{1pt}

        \textbf{Mistral} &
        \colorcell{51} & \colorcell{55} & \colorcell{40} & \colorcell{50} 
        \\\Xhline{1pt}

    \end{tabular}
    \caption{\textbf{Zero-Shot Performance across Humor Datasets.} Accuracy (\%) of LLaMA-2 and Mistral evaluated on the validation set of each dataset. Both models show limited humor detection ability in a zero-shot setting. \textit{Models used are the base versions without instruction tuning.}}

    \label{tab:zero shot results}
\end{table*}

\section{GPT-4-Turbo Prompt for Generating Dad Jokes}
\label{app:gpt-prompt}

    \#\#\# \\
    1. input: `A grizzly kept talking to me and annoyed me He was unbearable' \\
    output: `A grizzly kept talking to me and annoyed me He was intolerable' \\
    2. input: `For Christmas, I requested my family not to give me duplicates of the same item. Now I anticipate  
    receiving the missing sock next time.'  \\
    output: `For Christmas, I requested my family not to give me duplicates of the same item. Now I anticipate 
    receiving the other book next time.'  \\
    3. input: `My son’s fourth birthday was today, but when he came to see me I didn’t recognize him at first. I’d  
    never seen him be 4.'  \\
    output: `My son’s fourth birthday was today, but when he came to see me I didn’t recognize him at first. He grew
    up so fast.' \\
    4. input: `I asked my friend if he liked Nickleback. He told me that he never gave me any money' \\
    output: `I asked my friend if he liked Nickleback. He told me that he prefers Kings of Leon.' \\
    5. input: `I went to a bookstore and asked where the self-help section was The clerk said that if she told me,
    it would defeat the purpose.' \\
    output: `I went to a bookstore and asked where the self-help section was The clerk said it was in the third 
    aisle .' \\
    \#\#\# \\
    Using the examples in \#\#\# markers, please change some of the words in the following sentences to make 
    them non humorous. You can change anything but please change the least you can:

\section{Instruction Fine-Tuning Prompt}
\label{app:instruction-finetune-prompt}
Below is an instruction that describes a sentiment analysis task.\\\\ 
                            \#\#\# Instruction:\\  Given the following text, please determine if it should be classified as funny or not funny. Base your classification on humor elements such as wit, irony, absurdity, or comedic timing.
                            \\\\\#\#\# Input:
                            \\\{SAMPLE-TEXT\}\\\\
                            \#\#\# Response:\\
                            \{Yes/No\}\\

\section{Training and Hyperparameter Selection Details}
\label{app:training-details}

We conducted a systematic hyperparameter search using 4-fold cross-validation on each dataset. The search space was defined as the Cartesian product of the following values: learning rate \{3e-4, 5e-5, 6e-5\}, LoRA rank \{32, 64, 128\}, and LoRA alpha \{8, 16, 32\}, with a fixed seed of 42—resulting in 27 total combinations. To reduce computation time, we randomly sampled 10 configurations per dataset using random search. Each model was trained for 2 epochs.

The batch size was set to 4 when possible, and reduced to 2 for datasets with longer input sequences to fit within GPU memory limits. For each dataset, we selected the top 3 configurations based on median cross-validation accuracy across all evaluation datasets.

These configurations were used in the main experiments as follows:
\begin{itemize}
    \item \textbf{Single Dataset Training (A):} Each of the top three configurations was trained with four random seeds (7, 18, 28, 42). The configuration with the highest median accuracy across all datasets was selected for final evaluation on test set.
    \item \textbf{Double Dataset Training (B):} The top three configurations from each dataset in the pair (six total) were each trained with four random seeds. The best-performing configuration was selected based on median accuracy.
    \item \textbf{Triple Dataset Training (C):} For each held-out dataset, we used the top 3 configurations from each of the three training datasets (nine total). Each configuration was trained with four random seeds, again selecting the configuration with the highest median accuracy.
\end{itemize}
For all experimental setups, we report the mean accuracy across four different training seeds.

In the training process we used LoRA \citep{hu2022lora} for efficiency.
All models were trained on 5,000 examples (see Section \ref{sec:data}).
We ran all experiments using the Hugging Face \texttt{transformers} \citep{wolf2020transformers} and \texttt{peft} \citep{peft} libraries. Training was performed on a mix of \texttt{A6000}, \texttt{A40}, and \texttt{L40S} GPUs. Training and evaluation of all experiments took approximately 14 days in total.

\subsection{Best-found Hyperparameters}
We report the best training hyperparameters used for each model, selected based on highest median accuracy across evaluations. The parameters are listed in the following order: learning rate, LoRA rank, and LoRA alpha.

\begin{compactitem}

\item Mistral (Amazon): 0.0003, 64, 32
\item Mistral (Dad Jokes): 6.00E-05, 64, 8
\item Mistral (Headlines): 6.00E-05, 64, 32
\item Mistral (One Liners): 6.00E-05, 64, 32
\item Mistral (Amazon + Dad Jokes): 0.0003, 64, 8
\item Mistral (Amazon + Headlines):	0.0003, 64,	8
\item Mistral (Dad Jokes + Headlines): 6.00E-05, 64, 8
\item Mistral (Dad Jokes + One Liners): 0.0003, 128, 16
\item Mistral (Headlines + One Liners): 0.0003, 128, 16
\item Mistral (One Liners + Amazon): 0.0003, 64, 32
\item Mistral (Leave Out Amazon): 0.0003, 128, 16
\item Mistral (Leave Out Dad Jokes): 0.0003, 64, 32
\item Mistral (Leave Out Headlines): 0.0003, 64, 16
\item Mistral (Leave Out One Liners): 0.0003, 64, 32

\item LLaMA-2 (Amazon): 0.0003, 64, 8
\item LLaMA-2 (Dad Jokes): 0.0003, 128, 32 (Batch size = 4)
\item LLaMA-2 (Headlines): 6.00E-05, 128, 8
\item LLaMA-2 (One Liners): 5.00E-05, 32, 16 (Batch size = 4)
\item LLaMA-2 (Amazon + Dad Jokes): 0.0003, 32, 32 (Batch size = 4)
\item LLaMA-2 (Amazon + Headlines):	6.00E-05, 128,	32
\item LLaMA-2 (Dad Jokes + Headlines): 0.0003, 128, 32 (Batch size = 4)
\item LLaMA-2 (Dad Jokes + One Liners): 0.0003, 32, 32 (Batch size = 4)
\item LLaMA-2 (Headlines + One Liners): 0.0003, 32, 32 (Batch size = 4)
\item LLaMA-2 (One Liners + Amazon): 0.0003, 64, 8
\item LLaMA-2 (Leave Out Amazon): 0.0003, 32, 32 (Batch size = 4)
\item LLaMA-2 (Leave Out Dad Jokes): 0.0003, 32, 8
\item LLaMA-2 (Leave Out Headlines): 0.0003, 64, 8
\item LLaMA-2 (Leave Out One Liners): 0.0003, 64, 8

\end{compactitem}

\subsection{Packages and Configurations}

We used several widely adopted libraries for modeling, preprocessing, training, and evaluation. Below, we report the key packages and configurations we used:

    \paragraph{Transformers (Hugging Face)} We used pretrained models \texttt{mistralai/Mistral-7B-v0.1} and \texttt{meta-llama/Llama-2-7b-hf} via the \texttt{AutoModelForCausalLM} and \texttt{AutoTokenizer} interfaces. We set \texttt{tokenizer.pad\_token = tokenizer.eos\_token}. Models were loaded using 4-bit quantization via the \texttt{BitsAndBytesConfig} class, with the following settings: 
    \begin{compactitem}
        \item\texttt{load\_in\_4bit = True}
        \item\texttt{bnb\_4bit\_quant\_type = ``nf4''}
        \item\texttt{bnb\_4bit\_compute\_dtype = torch.float16}
    \end{compactitem}

    For generation-based evaluation, we used \texttt{model.generate()} with \texttt{max\_new\_tokens = 5}. 

    \paragraph{PEFT (LoRA)} We fine-tuned models using parameter-efficient fine-tuning via the \texttt{LoraConfig} class from the \texttt{peft} library. The following hyperparameters were used:
    \begin{compactitem}
        \item \texttt{lora\_dropout = 0.1}
        \item \texttt{bias = ``none''}
        \item \texttt{task\_type = ``CAUSAL\_LM''}
    \end{compactitem}

    \paragraph{TRL (SFTTrainer)} We trained models using the \texttt{SFTTrainer} class from the \texttt{trl} library. The training configuration was based on Hugging Face's \texttt{TrainingArguments}, with the following relevant settings:
    \begin{compactitem}
        \item \texttt{gradient\_accumulation\_steps = 4}
        \item \texttt{gradient\_checkpointing = True}
        \item \texttt{max\_seq\_length = 1024}
    \end{compactitem}

    \paragraph{Datasets} Dataset processing and construction were handled using the Hugging Face \texttt{datasets} library. For train/test splits, we used \texttt{train\_test\_split} with the following parameters:
    \begin{compactitem}
        \item \texttt{stratify\_by\_column = ``label''}
        \item \texttt{seed = 42}
    \end{compactitem}

    \paragraph{Scikit-learn} We used \texttt{StratifiedKFold} from \texttt{sklearn.model\_selection} for dataset partitioning. Cross-validation settings included:
    \begin{compactitem}
        \item \texttt{n\_splits = 4} 
        \item \texttt{shuffle = True}
        \item \texttt{random\_state = 1}
    \end{compactitem}
    For evaluation, we used the following metrics from \texttt{sklearn.metrics}: \texttt{accuracy\_score}, \texttt{precision\_score}, \texttt{recall\_score}, and \texttt{f1\_score}, all calculated with \texttt{pos\_label = 1}.

All relevant parameters, random seeds, and training configurations are documented in our code.

\section{Standard Deviations Across Seeds}
Table~\ref{tab:std-table} reports standard deviations across four training seeds for all experiments and test datasets.

\begin{table*}
    \centering
    \begin{tabular}{cccccc}
        \hline
        \multirow{2}{*}{\textbf{Train Dataset}} &
        \multirow{2}{*}{\textbf{Model}} &
        \multicolumn{4}{c}{\textbf{Test Dataset}} \\ \cline{3-6} &
                                            & \textbf{\textcolor{amazon-color}{Amazon}} & \textbf{\textcolor{dadjokes-color}{Dad Jokes}} & \textbf{\textcolor{headlines-color}{Headlines}} & \textbf{\textcolor{oneliners-color}{One Liners}} \\ \hline

        \multirow{2}{*}{\textbf{\textcolor{amazon-color}{Amazon}}} & LLaMA-2 &
        0.0017 & 0.0167 & 0.012 & 0.0141
        
        \\\cline{2-6} & Mistral &
        0.0068 & 0.0284 & 0.0137 & 0.008
        \\\Xhline{1pt}

        \multirow{2}{*}{\textbf{\textcolor{dadjokes-color}{Dad Jokes}}} & LLaMA-2 &
        0.0402 & 0.0046 & 0.0207 & 0.0147

        \\\cline{2-6} & Mistral &
        0.0137 & 0.0044 & 0.0124 & 0.0067 
        \\\Xhline{1pt}

        \multirow{2}{*}{\textbf{\textcolor{headlines-color}{Headlines}}} & LLaMA-2 &
        0.0047 & 0.0064 & 0.0045 & 0.0104

        \\\cline{2-6} & Mistral &
        0.0242 & 0.0114 & 0.0124 & 0.0067
        \\\Xhline{1pt}

        \multirow{2}{*}{\textbf{\textcolor{oneliners-color}{One Liners}}} & LLaMA-2 &
        0.0036 & 0.0088 & 0.0083 & 0.0045

        \\\cline{2-6} & Mistral & 
        0.0369 & 0.0034 & 0.0095 & 0.0026

        \\\Xhline{1pt}

        \multirow{2}{*}{\textbf{\textcolor{amazon-color}{Amazon} + \textcolor{dadjokes-color}{Dad Jokes}}} & LLaMA-2 &
        0.005 & 0.0076 & 0.011 & 0.0117

        \\\cline{2-6} & Mistral 
        & 0.0033 & 0.0045 & 0.0071 & 0.0237 
        \\\Xhline{1pt}

        \multirow{2}{*}{\textbf{\textcolor{amazon-color}{Amazon} + \textcolor{headlines-color}{Headlines}}} & LLaMA-2 &
        0.0034 & 0.0068 & 0.0025 & 0.0116
        
        \\\cline{2-6} & Mistral &
        0.0013 & 0.0308 & 0.0049 & 0.0168
        \\ \Xhline{1pt}

        \multirow{2}{*}{\textbf{\textcolor{dadjokes-color}{Dad Jokes} + \textcolor{headlines-color}{Headlines}}} & LLaMA-2 &
        0.023 & 0.008 & 0.0053 & 0.0147
        
        \\\cline{2-6} & Mistral &
        0.0191 & 0.0026 & 0.0037 & 0.0135
        \\ \Xhline{1pt}

        \multirow{2}{*}{\textbf{\textcolor{dadjokes-color}{Dad Jokes} + \textcolor{oneliners-color}{One Liners}}} & LLaMA-2 &
        0.011 & 0.0052 & 0.0092 & 0.0037

        \\\cline{2-6} & Mistral &
        0.057 & 0.0036 & 0.0071 & 0.0016
        \\ \Xhline{1pt}

        \multirow{2}{*}{\textbf{\textcolor{headlines-color}{Headlines} + \textcolor{oneliners-color}{One Liners}}} & LLaMA-2 &
        0.015 & 0.0168 & 0.0049 & 0.0024
        
        \\\cline{2-6} & Mistral
        & 0.0358 & 0.0051 & 0.0029 & 0.0013  
        \\\Xhline{1pt}

        \multirow{2}{*}{\textbf{\textcolor{oneliners-color}{One Liners} + \textcolor{amazon-color}{Amazon}}} & LLaMA-2 &
        0.0036 & 0.0206 & 0.0184 & 0.0023
        
        \\\cline{2-6} & Mistral &
        0.0041 & 0.0028 & 0.008 & 0.0037
        \\ \Xhline{1pt}

        \multirow{2}{*}{\textbf{\textcolor{amazon-color}{Held Out Amazon}}} & LLaMA-2 &
        0.0123 & 0.005 & 0.0075 & 0.0043 
        
        \\\cline{2-6} & Mistral &
        0.0545 & 0.0041 & 0.0054 & 0.0065
        \\\Xhline{1pt}

        \multirow{2}{*}{\textbf{\textcolor{dadjokes-color}{Held Out Dad Jokes}}} & LLaMA-2 &
        0.0092 & 0.0211 & 0.0054 & 0.019
        
        \\\cline{2-6} & Mistral &
        0.0028 & 0.0138 & 0.0011 & 0.0029
        \\ \Xhline{1pt}

        \multirow{2}{*}{\textbf{\textcolor{headlines-color}{Held Out Headlines}}} & LLaMA-2 &
        0.0041 & 0.007 & 0.0138 & 0.0029
        
        \\\cline{2-6} & Mistral &
        0.0029 & 0.0017 & 0.0056 & 0.0021 
        \\ \Xhline{1pt}

        \multirow{2}{*}{\textbf{\textcolor{oneliners-color}{Held Out One Liners}}} & LLaMA-2 &
        0.012 & 0.0045 & 0.0074 & 0.0146
        
        \\\cline{2-6} & Mistral &
        0.0038 & 0.004 & 0.0015 & 0.0305

        \\ \Xhline{1pt}
        
    \end{tabular}
    \caption{Standard deviation of accuracy scores over four training seeds for all experimental setups and test datasets, reported separately for LLaMA-2 and Mistral}

    \label{tab:std-table}
\end{table*}

\section{Transfer Accuracy Differences}
\label{app:transfer-diffs}
Table~\ref{tab:transfer diffs} reports the change in mean transfer accuracy (in percentage points) between experimental setups for each model and target dataset.

    \begin{table*}
    \centering
    \begin{tabular}{ccccc}
        \hline
        \multirow{2}{*}{\textbf{Target Dataset}} &
        \multirow{2}{*}{\textbf{Model}} &
        \multicolumn{3}{c}{\textbf{Transfer Accuracy Change}} \\ \cline{3-5} &
                                            & \textbf{Single $\rightarrow$ Double} & \textbf{Double $\rightarrow$ Triple} & \textbf{Single $\rightarrow$ Triple}  \\ \hline
        \multirow{2}{*}{\textbf{\textcolor{amazon-color}{Amazon}}} & LLaMA-2 &
        $\uparrow 5.58$ & $\uparrow 2.34$ & $\uparrow 7.92$
        
        \\\cline{2-5} & Mistral &
        $\uparrow 2.27$ & $\downarrow -2.6$  & $\downarrow -0.33$ 
        \\\Xhline{1pt}

        \multirow{2}{*}{\textbf{\textcolor{dadjokes-color}{Dad Jokes}}} & LLaMA-2 &
        $\downarrow -3.77$ & $\downarrow -0.31$ & $\downarrow -4.08$
        
        \\\cline{2-5} & Mistral &
        $\uparrow 1.31$ & $\downarrow -2.34$  & $\downarrow -1.03$ 
        \\\Xhline{1pt}

        \multirow{2}{*}{\textbf{\textcolor{headlines-color}{Headlines}}} & LLaMA-2 &
        $\uparrow 6.23$ & $\uparrow 2.33$ & $\uparrow 8.56$
        
        \\\cline{2-5} & Mistral &
        $\uparrow 2.55$ & $\uparrow 0.83$  & $\uparrow 3.38$ 
        \\\Xhline{1pt}

        \multirow{2}{*}{\textbf{\textcolor{oneliners-color}{One Liners}}} & LLaMA-2 &
        $\uparrow 4.02$ & $\downarrow -0.21$ & $\uparrow 3.81$
        
        \\\cline{2-5} & Mistral &
        $\uparrow 1.96$ & $\uparrow 3.54$  & $\uparrow 5.50$
        \\\Xhline{1pt}

        \multirow{2}{*}{\textbf{Average}} & LLaMA-2 &
        $\uparrow 3.02$ & $\uparrow 1.04$ & $\uparrow 4.05$
        
        \\\cline{2-5} & Mistral &
        $\uparrow 2.02$ & $\downarrow -0.14$  & $\uparrow 1.88$

        \\\hline
        
    \end{tabular}
    \caption{\textbf{Transfer Accuracy Differences Across Experiments.} Reported values reflect the change in mean transfer accuracy (in percentage points) between training setups for each model and target dataset. Accuracy is averaged over all relevant source datasets for each target. ↑ indicates improvement, and ↓ indicates degradation. For example, Mistral on Amazon improves from Single to Double training by 2.27 points, but decreases from Double to Triple training by 2.60 points. The “Average” row reports the mean change across all target datasets. Overall, LLaMA-2 tends to benefit more from increased training diversity than Mistral.}

    \label{tab:transfer diffs}
\end{table*}

\section{In-Domain Accuracy across Experiments}
\label{app:self-accuracy-plot}
\begin{figure*}
\begin{center}
  \includegraphics[width=1\textwidth]{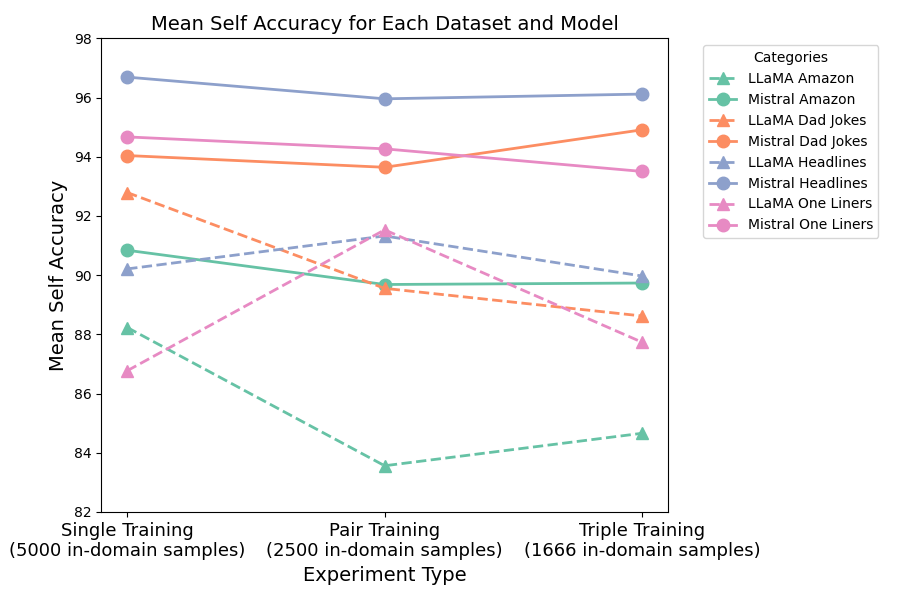}
  \caption{\textbf{Comparing self accuracy across the experiments.} Mistral results are shown with solid lines, LLaMA-2 with dotted lines. Colors represent different test datasets. The x-axis indicates the experiment type, along with the number of in-domain training samples, and the y-axis shows the mean accuracy on the in-domain dataset.  Mistral exhibits robust performance even as in-domain data decreases. LLaMA-2 results are less stable but show only minor decreases in accuracy.}

  \label{fig:NEW_self_accuracy_comparison}
  \end{center}
\end{figure*}

Figure~\ref{fig:NEW_self_accuracy_comparison} illustrates the mean in-domain accuracy across experiments for each model and dataset.

\section{Dataset Embeddings Similarity}
\label{app:cosine_heatmaps}
\begin{figure*}
\begin{center}
  \includegraphics[width=0.7\textwidth]{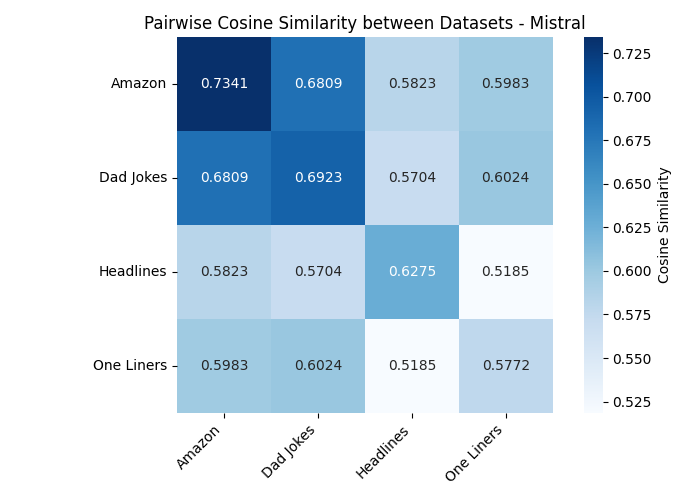}
  \caption[width=0.7\textwidth]{Mistral Embeddings Cosine Similarity Heatmap.}
  \label{fig:mistral_cosine_similarity}
  \end{center}
\end{figure*}

Figure~\ref{fig:mistral_cosine_similarity} shows the pairwise cosine similarity between datasets based on Mistral embeddings.

\section{Use of AI Assistance}
During the preparation of this work, the authors used ChatGPT, GitHub Copilot, and Google Gemini to assist with writing and coding. All content generated with these tools was reviewed and edited by the authors, who take full responsibility for the final publication.

\end{document}